\documentclass[lettersize,journal]{IEEEtran}
\usepackage{amsmath,amsfonts}
\usepackage{algorithmic}
\usepackage{algorithm}
\usepackage{array}
\usepackage[caption=false,font=normalsize,labelfont=sf,textfont=sf]{subfig}
\usepackage{textcomp}
\usepackage{stfloats}
\usepackage{url}

\usepackage{algorithm}
\usepackage{algorithmic}
\usepackage{amsmath} 

\usepackage{subcaption}
\usepackage{booktabs}

\usepackage{multirow}   
\usepackage{tabularx}   

\usepackage{verbatim}
\usepackage{graphicx}
\usepackage{cite}
\usepackage[hidelinks]{hyperref}
\begin{document}

\title{OHP-RL: Online Human Preference as Guidance in Reinforcement Learning for Robot Manipulation}

\author{
Yunyang Mo$^{1*}$, 
Jian Li$^{1*}$, 
Qiwei Wu$^{1}$,
Yihang Kang$^{1}$,
Renjing Xu$^{1\dagger}$,\\
$^{1}$The Hong Kong University of Science and Technology (Guangzhou) \\
\thanks{$^*$These authors contributed equally.}
\thanks{$\dagger$ Corresponding author}
}




\maketitle

\begin{abstract}
While reinforcement learning (RL) enables robots to acquire skills autonomously, its real-world deployment is severely limited by inefficient and unsafe exploration. Human-in-the-loop interventions offer a practical solution, yet existing methods typically exploit these interventions as auxiliary training signals, without fully capturing the richer information they provide about when and how autonomy should be guided.

Human interventions often encode relative preferences over behavior under safety and task constraints, rather than prescribing exact actions to imitate. Motivated by this perspective, we propose Online Human Preference as Guidance in Reinforcement
Learning (OHP-RL), a framework that leverages human interventions as preference information to guide policy learning. OHP-RL introduces a state-dependent preference gate that adaptively regulates when and to what extent human interventions should shape policy learning. This design enables the agent to benefit from intermittent and imperfect human feedback while preserving autonomous exploration and stable policy optimization.

We evaluate OHP-RL on three challenging real-world contact-rich manipulation tasks on a Franka robot. Across all tasks, OHP-RL consistently achieves strong success rates, faster convergence, and substantially lower human intervention effort than prior approaches. Moreover, the learned policies exhibit more stable and human-aligned behavior throughout training.
\end{abstract}


\section{Introduction}
\IEEEPARstart{A}{chieving} robust performance in real-world robotics remains a fundamental challenge, particularly in manipulation tasks requiring high-dimensional continuous control. Recent advances in reinforcement learning (RL) have demonstrated remarkable success in simulation~\cite{schulman2017proximal}, enabling agents to acquire complex manipulation skills through large scale trial-and-error training. However, transferring these successes directly to real-world robots remains difficult.

In physical environments, reward functions are often difficult to engineer, leading to sparse or poorly shaped learning signals. Real-world data collection is also costly and requires significant time. These challenges highlight the need for sample-efficient learning frameworks that can make effective use of limited real-world interaction data~\cite{Luo2024SERL,sample_eff1,sample_eff2}.

To improve sample efficiency and reduce the risks of unsafe exploration, human-in-the-loop reinforcement learning has emerged as a promising paradigm for real-world robotic training~\cite{luo2025precise}. Human guidance is especially valuable in robotic manipulation, where unsafe exploration can lead to unstable contacts, hardware stress, or irreversible failure states.

In this setting, human feedback often takes the form of corrective interventions when the robot exhibits undesirable behavior. Such interventions do more than prevent immediate failures: they also reveal which behaviors are preferred under task and safety constraints. This observation motivates viewing human-in-the-loop robotic RL through the lens of preference guidance, in which interventions are interpreted not merely as corrective actions, but as structured feedback for shaping policy learning~\cite{prefer1}.

We argue that these preference signals are particularly important for safe exploration in real-world robotics. In practice, safe behavior often requires adherence to strategies with human preference rather than aggressive reward maximization shortcuts. Compared with scalar value estimates alone, preference feedback can provide more informative guidance when value gradients are weak or ambiguous, leading to more effective policy updates and improved learning efficiency.

\begin{figure}[th]
    \centering
    \includegraphics[width=0.83\linewidth]{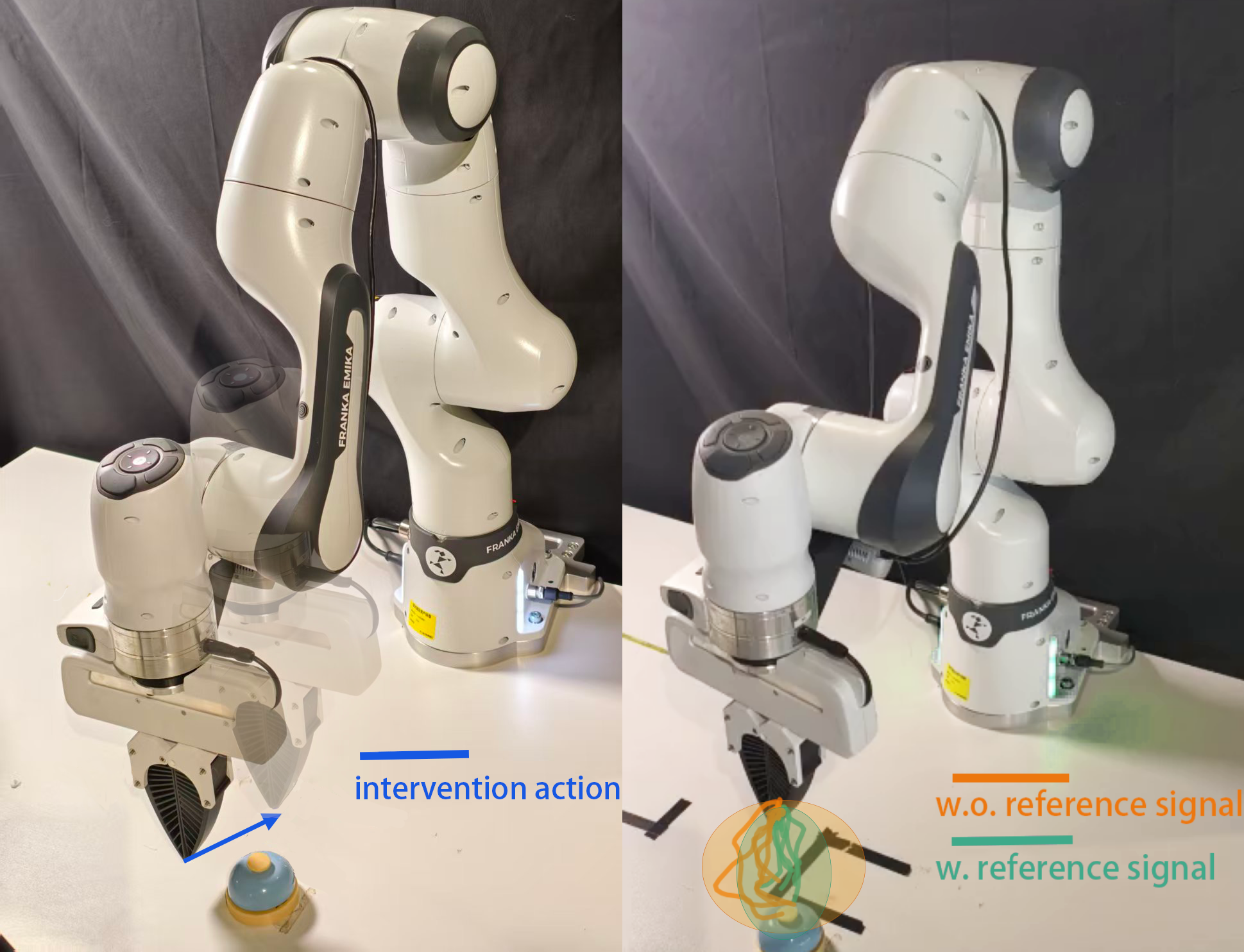}
    \caption{\textbf{Impact of preference signals on learning outcomes.} (Left) Physical intervention setup. (Right) Real-world execution trajectories of policies trained \textbf{with} (green) and \textbf{without} (brown) preference modeling.}
    \label{fig:intro}
\end{figure}

As illustrated in Fig.~\ref{fig:intro}, a real-world bell pressing task requires the robot to approach the button from above and press it safely. During training, when the end effector deviates toward unsafe configurations, a human operator intervenes to restore a valid pose, thereby providing implicit preference guidance for safety. The right panel compares execution trajectories of policies trained with and without preference modeling. Without such preference information, the policy tends to optimize for rapid contact and often exhibits undesirable lateral motions near the pressing surface. This example highlights the importance of exploiting human intervention signals more effectively so that RL agents learn behaviors that are safer and better aligned with human guidance.

Motivated by this perspective, our contributions are three-fold:
\begin{itemize}
    \item We propose OHP-RL, a human-in-the-loop reinforcement learning framework for sparse-reward real-world robot manipulation that addresses inefficient and unsafe exploration by interpreting human interventions as online preference signals.

    \item We evaluate OHP-RL on three challenging real-world manipulation tasks on a Franka robot. The results show that OHP-RL improves learning efficiency, reduces human intervention effort, and is more effectively transferred to autonomous execution.

    \item We conduct ablation studies and empirical analyses to validate the key design choices of OHP-RL. In particular, the results support both the adaptive preference target and the robustness of the preference guidance formulation under coarse or imprecise intervention targets.
\end{itemize}

\section{Related Work}

\subsection{Human-in-the-loop Learning}

Human-in-the-loop learning has been widely studied as a practical approach to improving robustness and safety in robot learning~\cite{HITP1, HITP2}. A representative line of work is interactive imitation learning, where human interventions are provided only when the learner visits problematic or out-of-distribution states. 
For instance, HG-DAgger~\cite{hg_dagger} improves over standard DAgger~\cite{dagger} by allowing the human to take over control during unsafe execution, reducing expert burden while collecting corrective data in critical regions. 
This idea has motivated subsequent work on data generation based on interventions~\cite{hoque2024intervengen}, expert feedback optimizing query efficiency or confidence awareness~\cite{hg_dagger_confidence}, and richer corrective supervision mechanisms~\cite{hg_dagger_pos_neg_constrains}.

Despite their practical success, intervention methods relying on imitation remain fundamentally tied to supervised correction. As a result, their performance is limited by the quality and coverage of intervention data, while the corrections need not be globally optimal. This has motivated more recent efforts to integrate human intervention directly into online reinforcement learning, allowing agents to continue improving beyond the intervention behavior while maintaining safety during exploration. 
For example, \cite{luo2025precise} combines human corrections with an RL pipeline that operates off-policy and maintains a high ratio of updates to data \cite{rlpd}, enabling efficient robotic learning in the physical world. However, intervention data in such systems are still primarily exploited as replay data.
Recent work has recognized that human interventions can be suboptimal~\cite{sil}, but still incorporates them mainly through partial imitation or behavioral regularization, which weakens the influence of corrective feedback on policy improvement. In contrast, our work treats human interventions as reference signals that directly shape policy updates, rather than as demonstrations or replay samples.

\subsection{Preference Guidance RL}

Preference guidance reinforcement learning trains policies from preference comparisons, providing a form of comparative supervision that can guide policy learning without requiring explicit action targets~\cite{pbrl1}. Early methods typically infer a reward model from human preference labels under the Bradley--Terry (BT) framework and then optimize the policy using standard RL algorithms~\cite{prefer1,bradley1952rank}, while more recent approaches optimize policies directly from preference comparisons~\cite{DPO}. 

In robotic settings where environment rewards are available, a key challenge is how to reconcile the policy improvement induced by human comparison signals with that driven by the underlying task reward. Existing methods such as HACO~\cite{haco} and PVP~\cite{pvp} incorporate comparison information derived from interventions primarily through critic learning. HACO learns from human intervention without explicitly using the environment reward, whereas PVP encodes intervention comparisons into a value function and propagates them through value updates~\cite{watkins1992q}. As a result, these methods do not explicitly regulate how comparison signals interact with environment rewards in shaping the actor, which can weaken or distort the corrective preference information during policy optimization. In contrast, our method utilizes preference signals obtained through interventions to guide policy updates more directly.

Unlike many preference guidance formulations based on comparisons of full trajectories or segments~\cite{prefer1,chen2022human}, our interactive setting provides online preference information exactly when an intervention occurs, making it more suitable for online policy improvement.

\section{Problem Formulation}

We model the real-world robotic task as a partially observable Markov decision process (POMDP), where the full environment state is not directly accessible. Instead, the robot receives observations consisting of multi-view visual features and proprioceptive robot states. In practice, we use these observations as a compact state representation and approximate the problem as an MDP
$\mathcal{M} = (\mathcal{S}, \mathcal{A}, P, r, \gamma)$,
where the transition dynamics $P$ are unknown and can only be accessed through physical interaction. The task is specified by a sparse success reward, with $r(s_t,a_t)=1$ upon task completion and $0$ otherwise. Our objective is to learn a policy $\pi$ that maximizes the expected discounted return over an episode horizon $T$:
\[
\mathbb{E}_{\pi}\!\left[\sum_{t=0}^{T-1}\gamma^{t} r(s_t,a_t)\right].
\]
This sparse-reward, partially observable setting makes exploration particularly challenging in real-world robotic learning.

\begin{figure*}[th]
    \centering
    \includegraphics[width=0.95\textwidth]{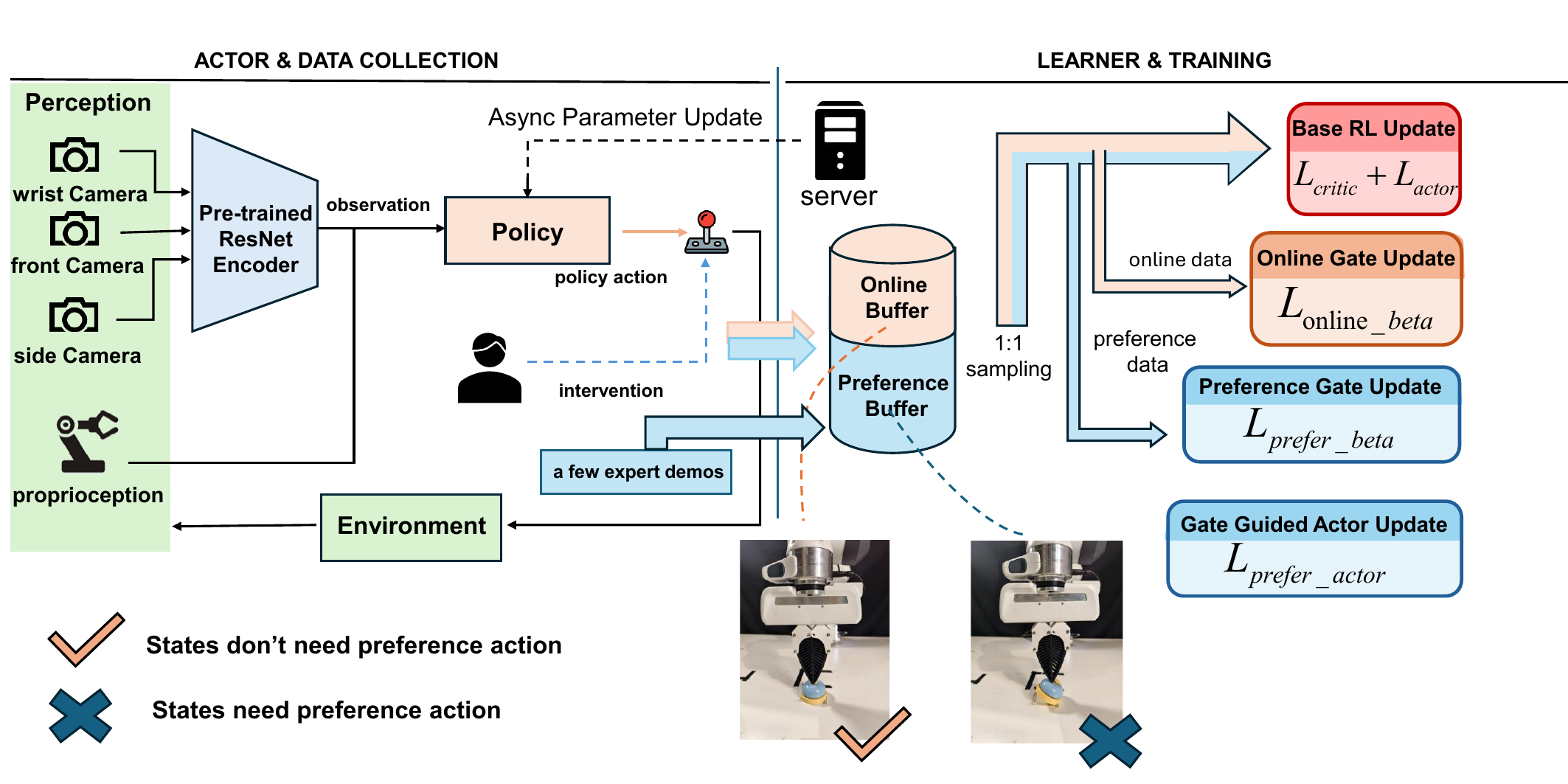}
    \caption{Overview of OHP-RL with online interaction and preference guidance optimization.}
    \label{fig:framework}
\end{figure*}

\section{Method}

We present Online Human Preference as Guidance in Reinforcement Learning (OHP-RL), an asynchronous actor--critic framework that integrates human intervention as online preference supervision under a high update-to-data (UTD) ratio~\cite{Luo2024SERL}. As illustrated in Fig.~\ref{fig:framework}, OHP-RL consists of four update modules: (i) \textbf{Base RL Update}, which learns the critic and actor from actual environment transitions; (ii) \textbf{Online Gate Update}, which regularizes the preference gate on purely online states; (iii) \textbf{Preference Gate Update}, which fits the gate to preference targets obtained through interventions; and (iv) \textbf{Gate Guided Actor Update}, which refines the policy using gated pairwise preference supervision. 

\subsection{Human Intervention and Data Collection}
\label{method:collect}

OHP-RL maintains two replay buffers: a standard online replay buffer $\mathcal{D}_{\text{online}}$ and a preference buffer $\mathcal{D}_{p}$. 

During online interaction, the actor proposes an action $a_t \sim \pi_\phi(\cdot|s_t)$. If no intervention occurs, the executed transition 
\[
(s_t,a_t,r_t,d_t,s_{t+1})
\]
is stored in $\mathcal{D}_{\text{online}}$. If a human override occurs, the human action is recorded as the preferred action $a_p$, while the policy action proposed at the same state is recorded as the weak action $a_w$. If the policy is uninitialized, $a_w$ is drawn from a uniform action prior. The resulting intervention tuple
\[
(s_t,a_p,a_w,r_t,d_t,s_{t+1})
\]
is stored in $\mathcal{D}_{p}$.

The distinction between $a_p$ and $a_w$ serves two different purposes in later updates. The preferred action $a_p$ is the \emph{executed} action and is therefore used in the base RL update. By contrast, $a_w$ is the action explicitly rejected at the time of intervention and is therefore used as the unpreferred comparator in preference guidance actor learning.

Before online training begins, $\mathcal{D}_{\text{online}}$ is prefilled with a small amount of rollout data collected by the initial policy, whose actions are nearly random at this stage, while $\mathcal{D}_{p}$ is prefilled with a small set of expert demonstration trajectories that successfully complete the task. Both buffers are streamed asynchronously to a background learner, whose updated parameters are periodically synchronized back to the interacting actor.

\subsection{Base Reinforcement Learning Update}
\label{method:base}

At each learner step, we sample mini-batches $\mathcal{B}_{\text{online}} \sim \mathcal{D}_{\text{online}}$ and $\mathcal{B}_{p} \sim \mathcal{D}_{p}$ using a symmetric 1:1 ratio. Since tuples in $\mathcal{D}_{p}$ contain both $a_p$ and $a_w$, we extract only the executed preferred action $a_p$ to form standard RL transitions. This yields the base RL batch
\begin{equation}
\begin{aligned}
\mathcal{B}_{\text{base}}
=
\mathcal{B}_{\text{online}}
\cup
\Big\{
&(s_t,a_p,r_t,d_t,s_{t+1}) \\
&\mid (s_t,a_p,a_w,r_t,d_t,s_{t+1}) \in \mathcal{B}_{p}
\Big\}.
\end{aligned}
\end{equation}
The base RL update therefore consists of minimizing $L_{\text{critic}}$ and $L_{\text{actor}}$ on $\mathcal{B}_{\text{base}}$.

\paragraph{Critic Update.}
The critic parameters $\theta$ are optimized by minimizing the soft Bellman residual:
\begin{equation}
L_{\text{critic}}
=
\mathbb{E}_{(s_t,a_t,r_t,d_t,s_{t+1}) \sim \mathcal{B}_{\text{base}}}
\left[
\left(
Q_\theta(s_t,a_t)-y_t
\right)^2
\right],
\end{equation}
with target
\begin{equation}
y_t
=
r_t
+
\gamma(1-d_t)
\left(
Q_{\bar{\theta}}(s_{t+1},\tilde{a}_{t+1})
-
\alpha \log \pi_\phi(\tilde{a}_{t+1}\mid s_{t+1})
\right),
\end{equation}
where $\tilde{a}_{t+1}\sim\pi_\phi(\cdot\mid s_{t+1})$, $\bar{\theta}$ denotes the target critic parameters, and $\alpha$ is the entropy temperature.

\paragraph{Actor Update.}
The actor parameters $\phi$ are updated by minimizing
\begin{equation}
L_{\text{actor}}
=
\mathbb{E}_{s_t \sim \mathcal{B}_{\text{base}},\, \tilde{a}_t \sim \pi_\phi(\cdot\mid s_t)}
\left[
\alpha \log \pi_\phi(\tilde{a}_t\mid s_t)
-
Q_\theta(s_t,\tilde{a}_t)
\right].
\end{equation}
This base update learns from actual environment transitions only, and is therefore unaffected by the unexecuted weak actions stored in $\mathcal{D}_{p}$.

\subsection{Preference Gate Learning}
\label{method:prefer}

Because preference supervision is available only on intervention states in $\mathcal{D}_{p}$, directly regularizing the actor equally would risk extrapolating human guidance out of distribution. To address this, we introduce a state-dependent preference gate $\beta_\psi(s)\in(0,1)$.

\paragraph{Target Construction.}
For intervention states, the gate should take large values only when the action preferred by the human remains superior to the action currently induced by the policy. Accordingly, for each $s_t \in \mathcal{D}_{p}$, we sample a policy action
\[
\tilde{a}_w \sim \pi_\phi(\cdot\mid s_t),
\]
and define the advantage as
\begin{equation}
A(s_t)
=
Q_\theta(s_t,a_p)-Q_\theta(s_t,\tilde{a}_w).
\end{equation}
Here, $\tilde{a}_w$ is sampled from the current policy because the stored weak action $a_w$ may no longer reflect the policy's latest behavior. If the current policy already proposes an action whose critic value matches or exceeds that of the human action, the preference gate should be reduced accordingly. We further analyze this design choice in Section~\ref{ablation}.
 We therefore define the gate target as
\begin{equation}
\beta_{\text{target}}(s_t)
=
\sigma\!\left(A(s_t)\right),
\end{equation}
where $\sigma(\cdot)$ is the sigmoid function.

\paragraph{Online Gate Update.}
To prevent undesired preference extrapolation in states lacking explicit human supervision, we apply a conservative regularization to the gating mechanism:
\begin{equation}
L_{\text{online-}\beta}(\psi)
=
\mathbb{E}_{s_t \sim \mathcal{D}_{\text{online}}}
\left[
\beta_\psi(s_t)^2
\right].
\end{equation}

\paragraph{Preference Gate Update.}

For intervention states, we fit the gate to the target computed by the critic:
\begin{equation}
\label{eq:preference_gate}
L_{\text{prefer-}\beta}(\psi)
=
\mathbb{E}_{s_t \sim \mathcal{D}_{p}}
\left[
\left(
\beta_\psi(s_t)-\beta_{\text{target}}(s_t)
\right)^2
\right].
\end{equation}
These two objectives yield a conservative gate outside regions supported by interventions, while activating preference guidance only where human intervention remains advantageous under the current critic.

\begin{algorithm}[t]
\caption{Online Human Preference as Guidance in Reinforcement
Learning (OHP-RL)}
\label{alg:ohprl}
\begin{algorithmic}[1]
\REQUIRE Initial parameters $\phi,\theta,\psi$; buffers $\mathcal{D}_{\text{online}}, \mathcal{D}_{p}$
\STATE Prefill $\mathcal{D}_{\text{online}}$ and $\mathcal{D}_{p}$ with initial rollout data and successful expert demonstrations, respectively
\WHILE{training}
    \STATE Sample $a_t \sim \pi_\phi(\cdot \mid s_t)$
    \IF{human intervention occurs}
        \STATE Set $a_p \leftarrow a^H$, $a_w \leftarrow a_t$
        \STATE Execute $a_p$ and observe $(r_t,d_t,s_{t+1})$
        \STATE $\mathcal{D}_{p} \leftarrow \mathcal{D}_{p} \cup \{(s_t,a_p,a_w,r_t,d_t,s_{t+1})\}$
    \ELSE
        \STATE Execute $a_t$ and observe $(r_t,d_t,s_{t+1})$
        \STATE $\mathcal{D}_{\text{online}} \leftarrow \mathcal{D}_{\text{online}} \cup \{(s_t,a_t,r_t,d_t,s_{t+1})\}$
    \ENDIF

    \STATE Sample $\mathcal{B}_{\text{online}} \sim \mathcal{D}_{\text{online}}$ and $\mathcal{B}_{p} \sim \mathcal{D}_{p}$
    \STATE Construct $\mathcal{B}_{\text{base}}$ from $\mathcal{B}_{\text{online}}$ and preferred transitions in $\mathcal{B}_{p}$

    \STATE \textbf{1. Base RL Update}
    \STATE $\theta \leftarrow \theta - \eta_\theta \nabla_\theta L_{\text{critic}}(\mathcal{B}_{\text{base}})$
    \STATE $\phi \leftarrow \phi - \eta_\phi \nabla_\phi L_{\text{actor}}(\mathcal{B}_{\text{base}})$

    \STATE \textbf{2. Online Gate Update}
    \STATE $\psi \leftarrow \psi - \eta_\beta \nabla_\psi L_{\text{online-}\beta}(\mathcal{B}_{\text{online}})$

    \STATE \textbf{3. Preference Gate Update}
    \STATE Sample $\tilde a_w \sim \pi_\phi(\cdot \mid s_t)$ for $s_t \in \mathcal{B}_{p}$
    \STATE $\psi \leftarrow \psi - \eta_\beta \nabla_\psi L_{\text{prefer-}\beta}(\mathcal{B}_{p})$

    \STATE \textbf{4. Gate-Guided Actor Update}
    \STATE $\phi \leftarrow \phi - \eta_\phi \nabla_\phi \big(\lambda_{\text{pref}} L_{\text{prefer-actor}}(\mathcal{B}_{p})\big)$
\ENDWHILE
\end{algorithmic}
\end{algorithm}
\subsection{Preference Guidance Actor Update}
\label{method:actor}

We apply a pairwise preference update guided by the gate on tuples from $\mathcal{D}_{p}$. For each intervention state, the current policy action $\tilde a_t \sim \pi_\phi(\cdot \mid s_t)$ is encouraged to move toward the action preferred by the human $a_p$ and away from the rejected weak action $a_w$. The resulting actor loss is
\begin{equation}
\begin{aligned}
L_{\text{prefer-actor}}(\phi)
=
\mathbb{E}_{(s_t,a_p,a_w)\sim \mathcal{D}_{p}}
\Big[
\beta_\psi(s_t)
\big(
\|\tilde a_t-a_p\|_2 \\
\qquad\qquad\qquad\qquad
-
\|\tilde a_t-a_w\|_2
\big)
\Big],
\quad
\tilde a_t \sim \pi_\phi(\cdot \mid s_t).
\end{aligned}
\end{equation}

Overall, the actor is optimized by two complementary objectives: the base RL loss $L_{\text{actor}}$ and the gate-guided preference loss $L_{\text{prefer-actor}}$. We combine them as
\begin{equation}
L_{\text{actor-total}}
=
L_{\text{actor}}
+
\lambda_{\text{pref}} L_{\text{prefer-actor}},
\end{equation}
where $\lambda_{\text{pref}}$ is a global balancing weight. In practice, $\lambda_{\text{pref}}$ only controls the overall scale of the preference term, while the state-dependent preference strength is learned by $\beta_\psi(s_t)$, making the overall preference modulation less sensitive to this global hyperparameter.

\subsection{Algorithm Overview}
\label{method:overview}

Algorithm~\ref{alg:ohprl} summarizes the OHP-RL training loop and its four update components, corresponding to the framework in Fig.~\ref{fig:framework}.

\section{Experiments}
\label{experiments}
We evaluate OHP-RL through comprehensive real-world experiments, examining its task performance, learning efficiency, and the contribution of its key components through comparative and ablation studies.

\subsection{Experimental Setup}

All experiments are conducted on a Franka robotic arm. The robot is equipped with an end-effector-mounted D435i camera for ego-centric observations, together with a front-view L515 and a side-view D435 for external workspace perception. Human interventions are provided via a 3D SpaceMouse, enabling real-time Cartesian control overrides during online interaction.

For perception, we use an ImageNet-pretrained ResNet-10~\cite{resnet} as the visual encoder, followed by a two-layer MLP for latent feature processing and policy learning. Observations include multi-view RGB images and robot proprioceptive states, including end-effector pose, twist, force, and torque.

The policy outputs a 6D Cartesian delta pose relative to the current end-effector pose, which is executed by a customized Cartesian impedance controller following~\cite{Luo2024SERL}.

\begin{figure}[t]
    \centering
    \begin{minipage}[b]{0.32\columnwidth}
        \centering
        \includegraphics[width=\textwidth]{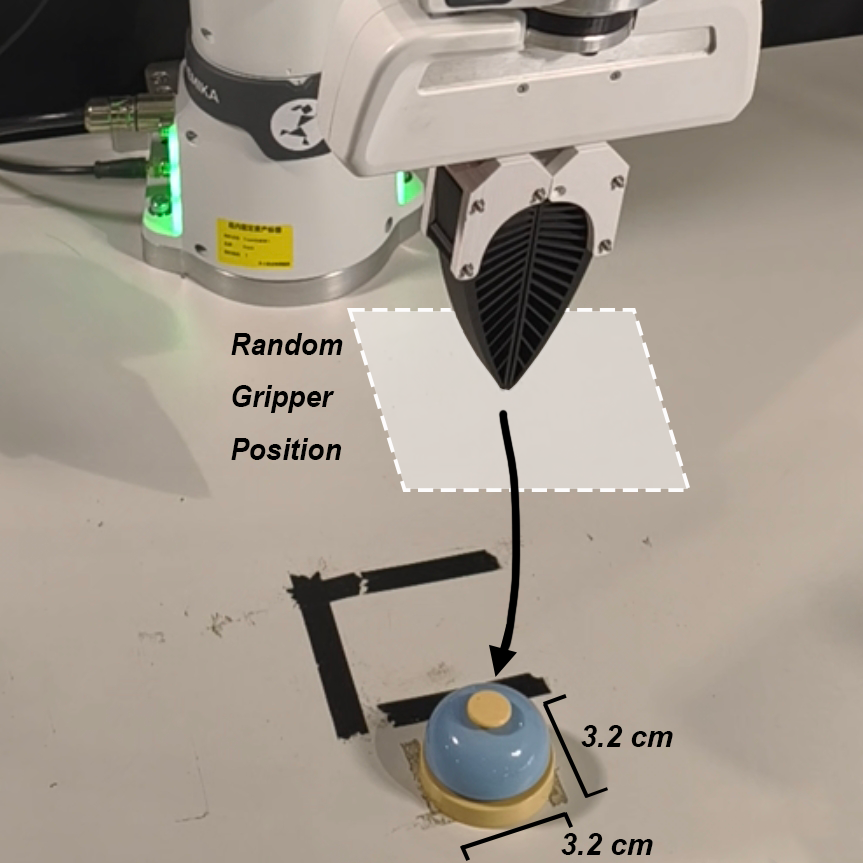}
        \caption*{(a) Press Bell}
    \end{minipage}\hfill
    \begin{minipage}[b]{0.32\columnwidth}
        \centering
        \includegraphics[width=\textwidth]{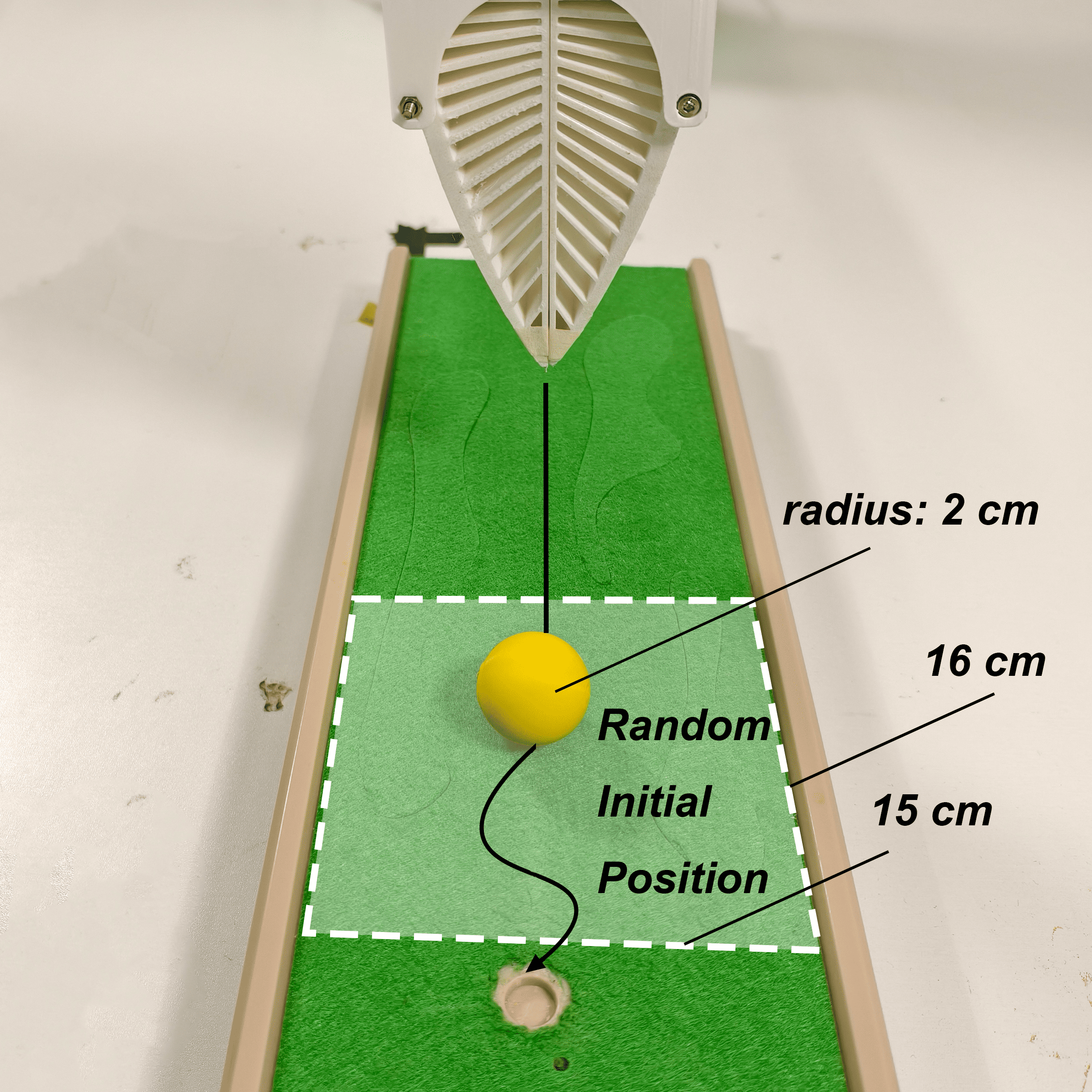}
        \caption*{(b) Push Ball}
    \end{minipage}\hfill
    \begin{minipage}[b]{0.32\columnwidth}
        \centering
        \includegraphics[width=\textwidth]{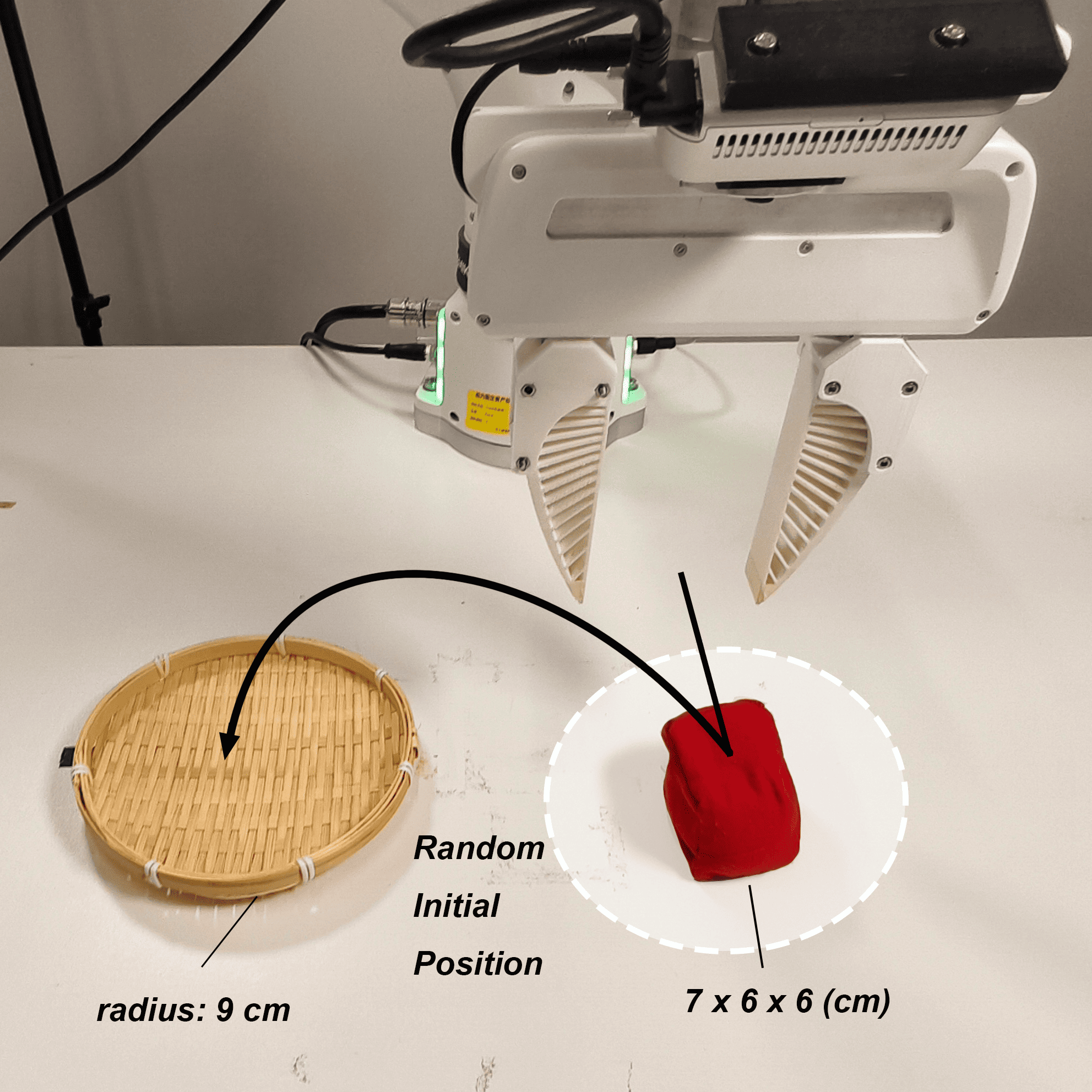}
        \caption*{(c) Move Sand Bag}
    \end{minipage}

    \caption{The three real-world manipulation tasks used in our experiments. All tasks are evaluated with sparse binary success rewards, while exhibiting different sources of difficulty in real-world interaction.}
    \label{fig:tasks_setup}
\end{figure}

\subsection{Task Definition}

We evaluate OHP-RL on three real-world manipulation tasks with sparse binary success rewards, as illustrated in Fig.~\ref{fig:tasks_setup}. In all tasks, the robot learns from online interaction with occasional human interventions, which are further converted into pairwise preference supervision. Task completion is determined by a binary reward classifier consisting of a pretrained ResNet-10 encoder and a two-layer MLP. A reward of $1$ is assigned upon successful completion and $0$ otherwise.

The three tasks are designed to form a progressive spectrum of real-world difficulty. \textit{Press Bell} is the simplest task in terms of task structure, but it already requires precise contact direction: the robot must approach the bell vertically rather than collide from the side and tip it over. \textit{Push Ball} is more challenging because successful completion depends on long-horizon interaction with uncertain contact geometry and surface friction, making the ball trajectory difficult to predict and control. \textit{Move Sand Bag} is the most difficult task, as the robot must manipulate a deformable sand bag whose shape and mass distribution change during grasping and transport, making both grasp acquisition and stable placement substantially harder. As shown in Fig.~\ref{fig:tasks_setup}, these three tasks therefore expose increasingly difficult forms of physical interaction, ranging from motion sensitive to contact, to object transport sensitive to friction, to the manipulation of deformable objects.

\subsection{Implementation Details}
All experiments are conducted in an asynchronous setup using two NVIDIA 16GB GPUs, with one GPU dedicated to the learner and the other to the actor. Unless otherwise specified, all fundamental network architectures, update-to-data ratios, training horizons, and hyperparameters are kept fixed across tasks.

\subsection{Baselines}

We compare OHP-RL against five representative baselines that use human intervention in different ways:

\begin{itemize}
    \item \textbf{HG-DAgger.} Pure imitation learning without reinforcement learning from environment rewards, where human interventions are treated as corrective demonstrations for behavior cloning.

    \item \textbf{HIL-SERL.} An interactive RL method that incorporates data from interventions primarily through value updates for the critic, operating off policy with a high ratio of updates to data.

    \item \textbf{SIL-RI.} This approach uses intervention data as supervision for an extra imitation actor, explicitly regularizing the policy toward the behavior induced by that actor.

    \item \textbf{HACO.} This preference guidance method captures human guidance entirely through value learning derived from interventions, without using environment rewards during policy optimization.

    \item \textbf{PVP*.} A modified variant of PVP in which intervention comparisons are propagated through critic learning. To better suit our setting with sparse rewards, we assign a smaller target critic value (0.1) to only the first transition of each consecutive intervention segment as its intervention value alignment, reducing critic distortion while preserving the original mechanism for propagating comparisons.
\end{itemize}

These baselines allow us to compare OHP-RL against alternative ways of using intervention and preference information.

\begin{figure*}[t]
    \centering
    \begin{minipage}[b]{0.32\textwidth}
        \centering
        \includegraphics[width=\textwidth]{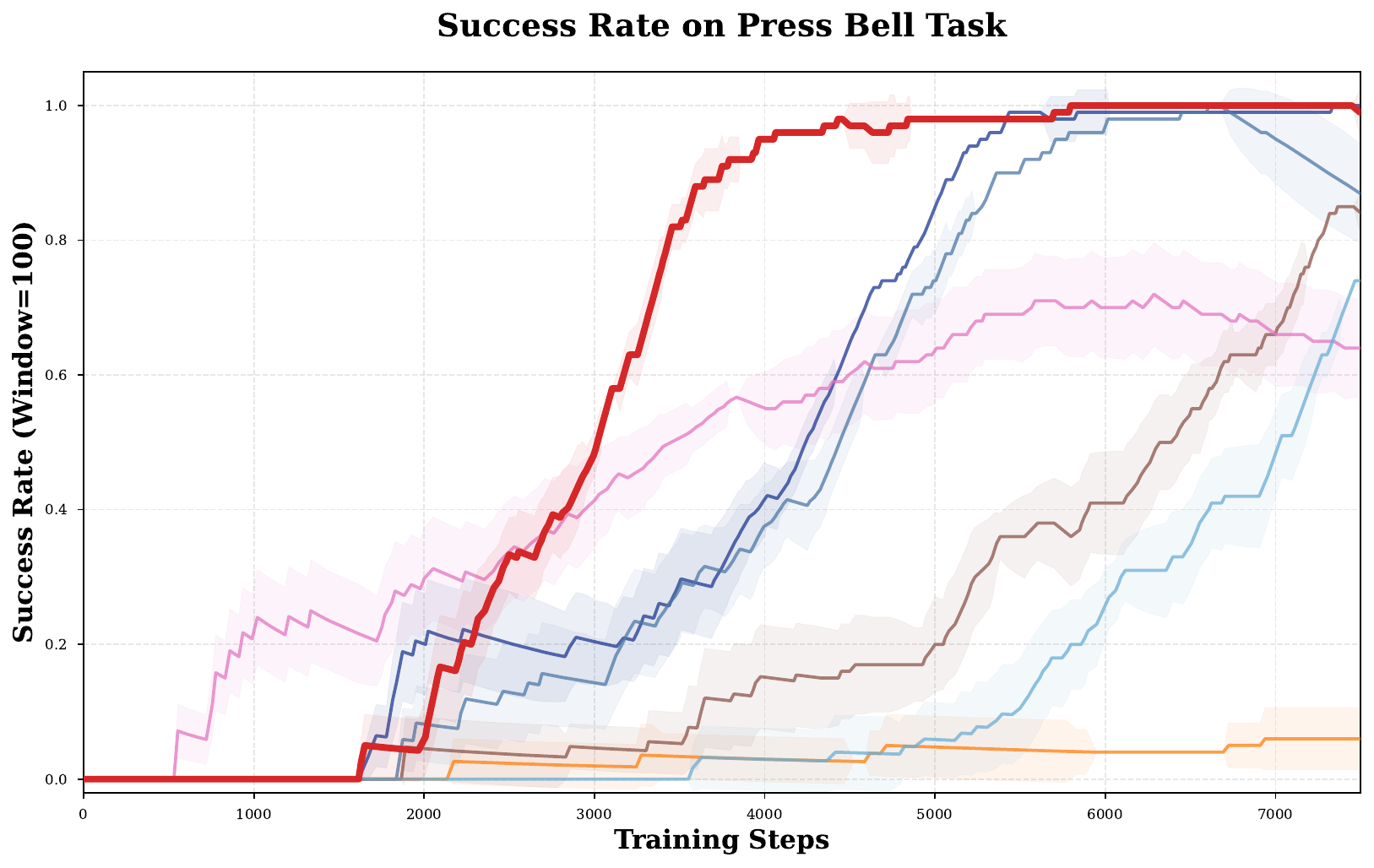}
        \caption*{(a) Success Rate}
    \end{minipage}
    \hfill
    \begin{minipage}[b]{0.32\textwidth}
        \centering
        \includegraphics[width=\textwidth]{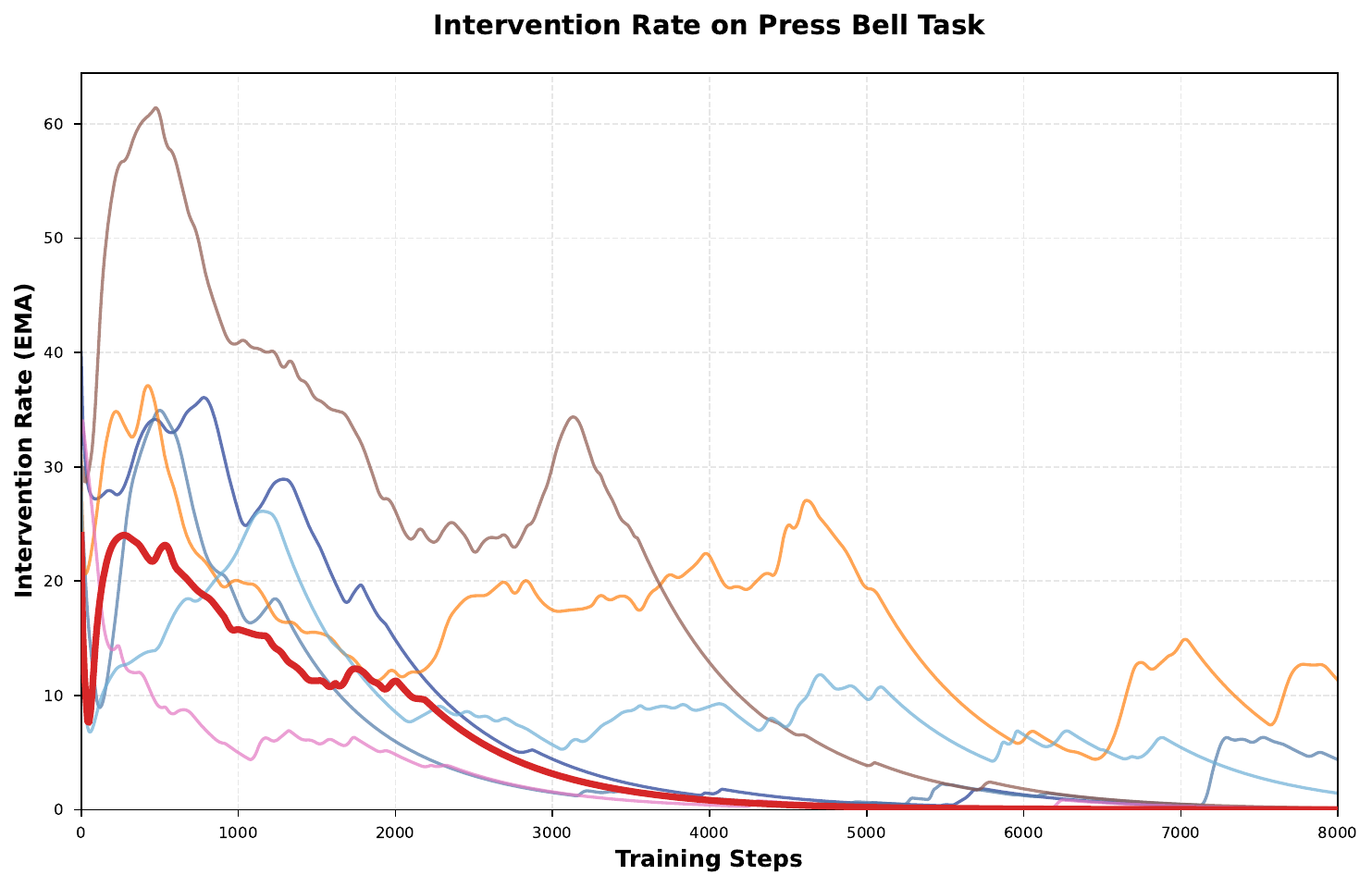}
        \caption*{(b) Intervention Rate}
    \end{minipage}
    \hfill
    \begin{minipage}[b]{0.32\textwidth}
        \centering
        \includegraphics[width=\textwidth]{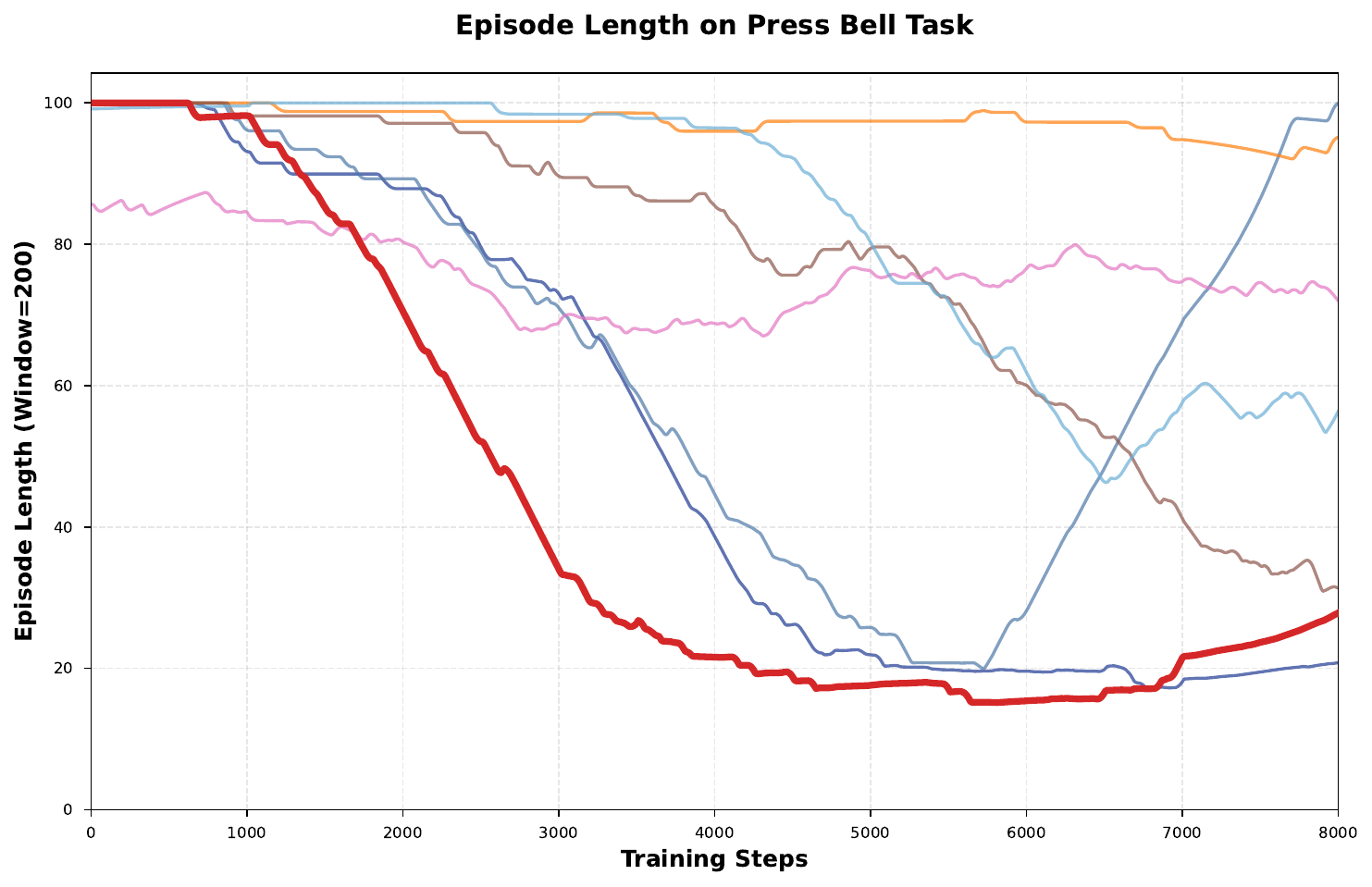}
        \caption*{(c) Episode Length}
    \end{minipage}

    \caption{Training performance on the \textit{Press Bell} task.}
    \label{fig:results_press_bell}
\end{figure*}

\begin{figure*}[t]
    \centering
    \begin{minipage}[b]{0.32\textwidth}
        \centering
        \includegraphics[width=\textwidth]{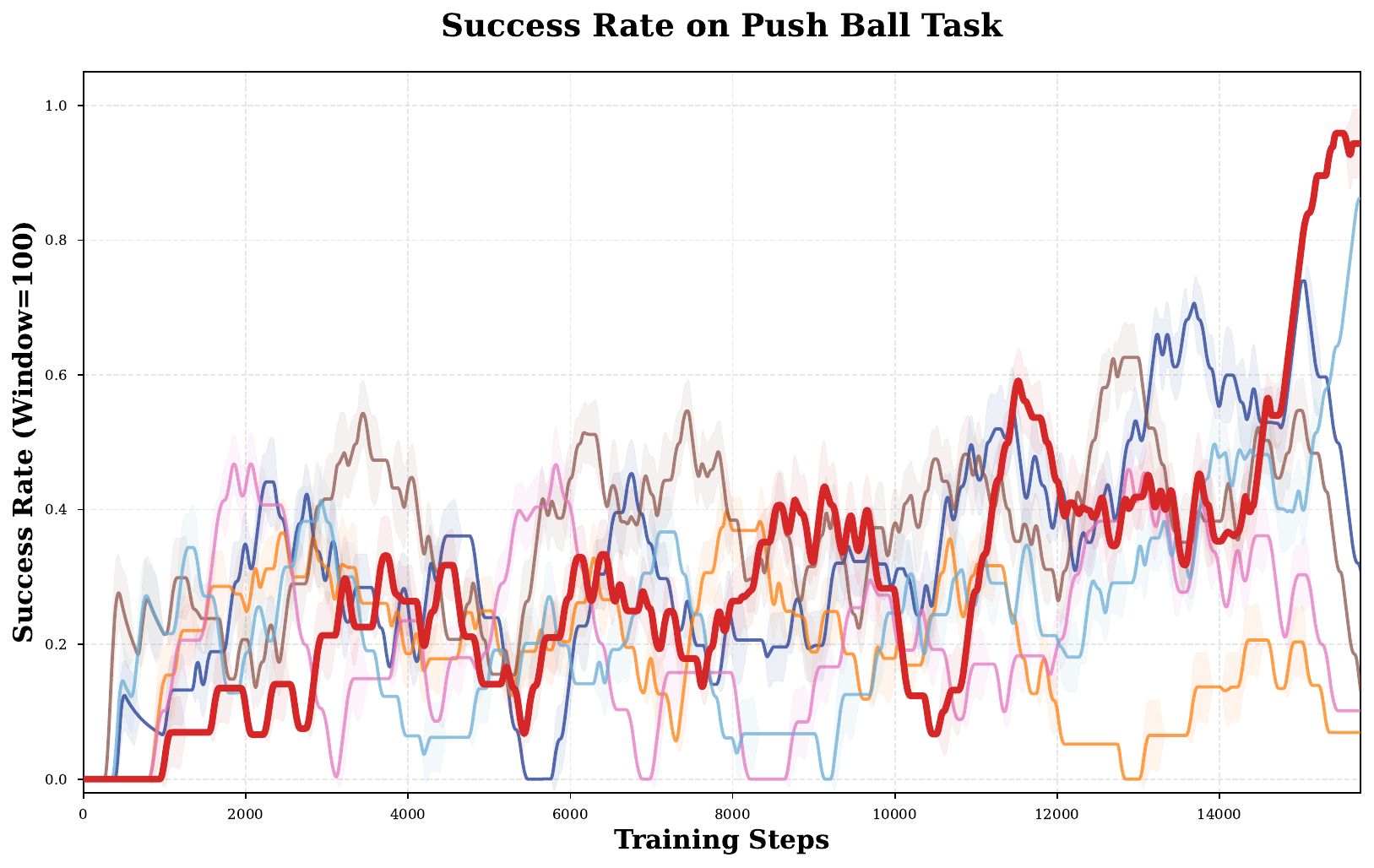}
        \caption*{(a) Success Rate}
    \end{minipage}
    \hfill
    \begin{minipage}[b]{0.32\textwidth}
        \centering
        \includegraphics[width=\textwidth]{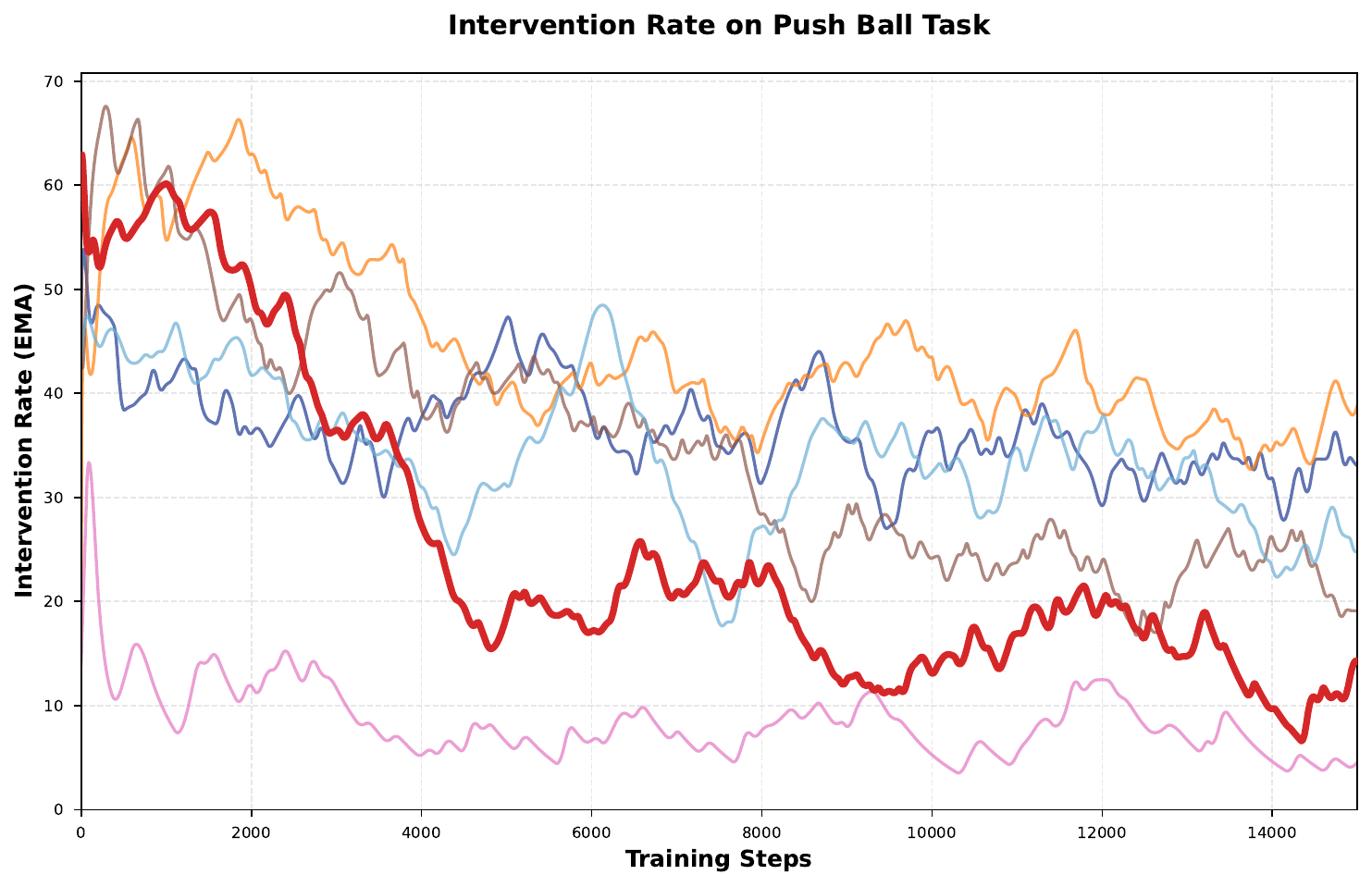}
        \caption*{(b) Intervention Rate}
    \end{minipage}
    \hfill
    \begin{minipage}[b]{0.32\textwidth}
        \centering
        \includegraphics[width=\textwidth]{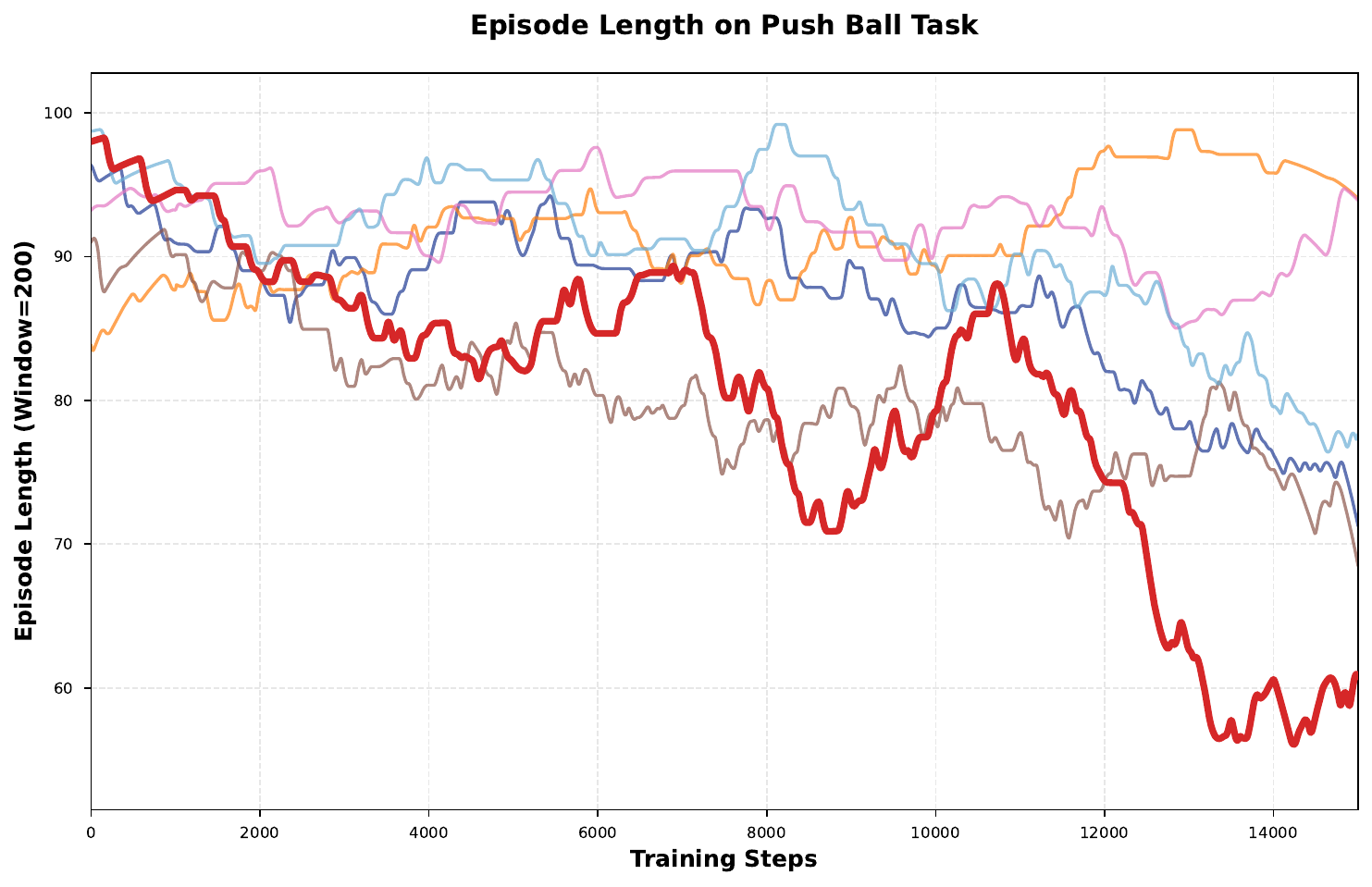}
        \caption*{(c) Episode Length}
    \end{minipage}

    \caption{Training performance on the \textit{Push Ball} task. }
    \label{fig:results_push_ball}
\end{figure*}

\begin{figure*}[t]
    \centering
    \begin{minipage}[b]{0.32\textwidth}
        \centering
        \includegraphics[width=\textwidth]{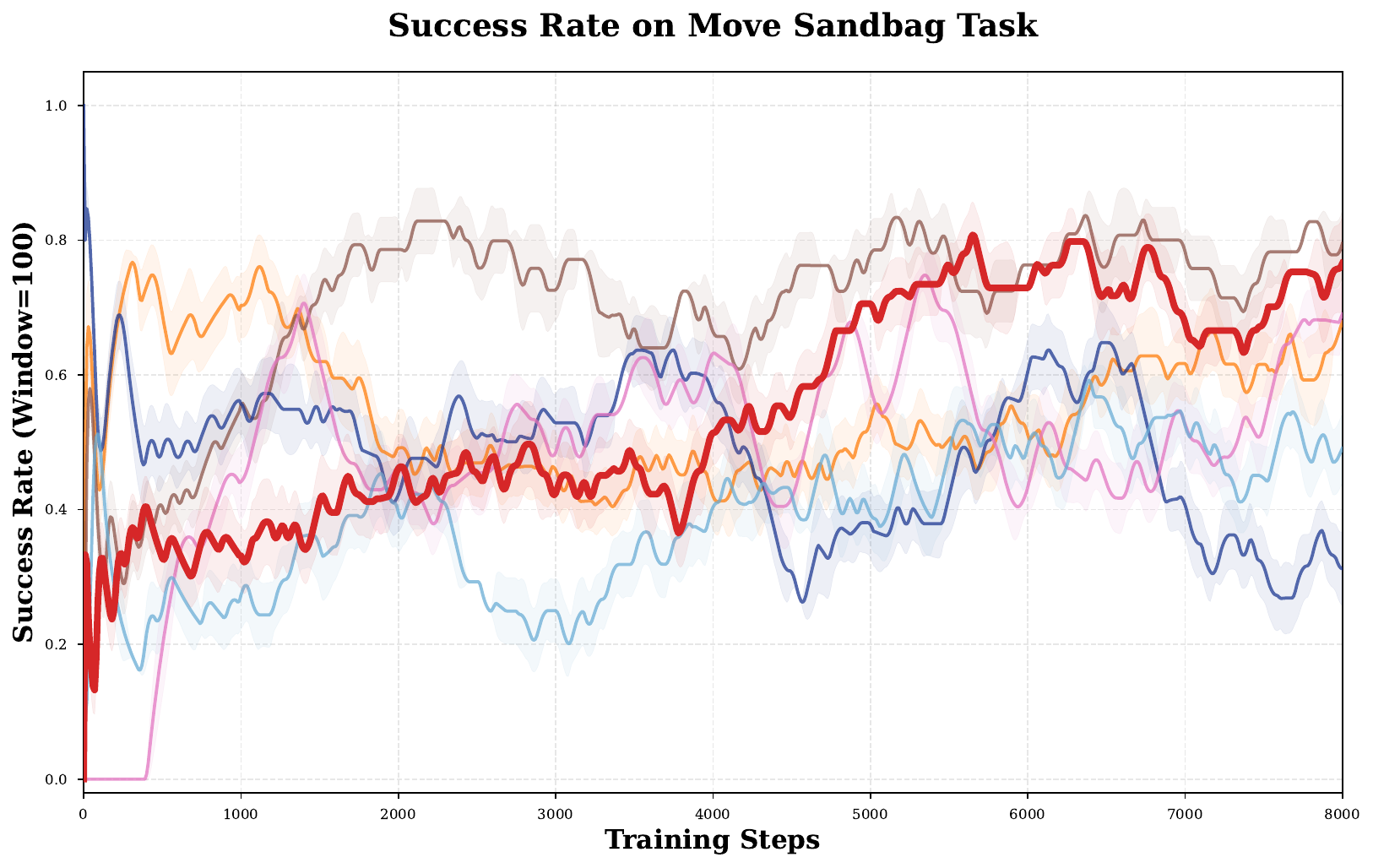}
        \caption*{(a) Success Rate}
    \end{minipage}
    \hfill
    \begin{minipage}[b]{0.32\textwidth}
        \centering
        \includegraphics[width=\textwidth]{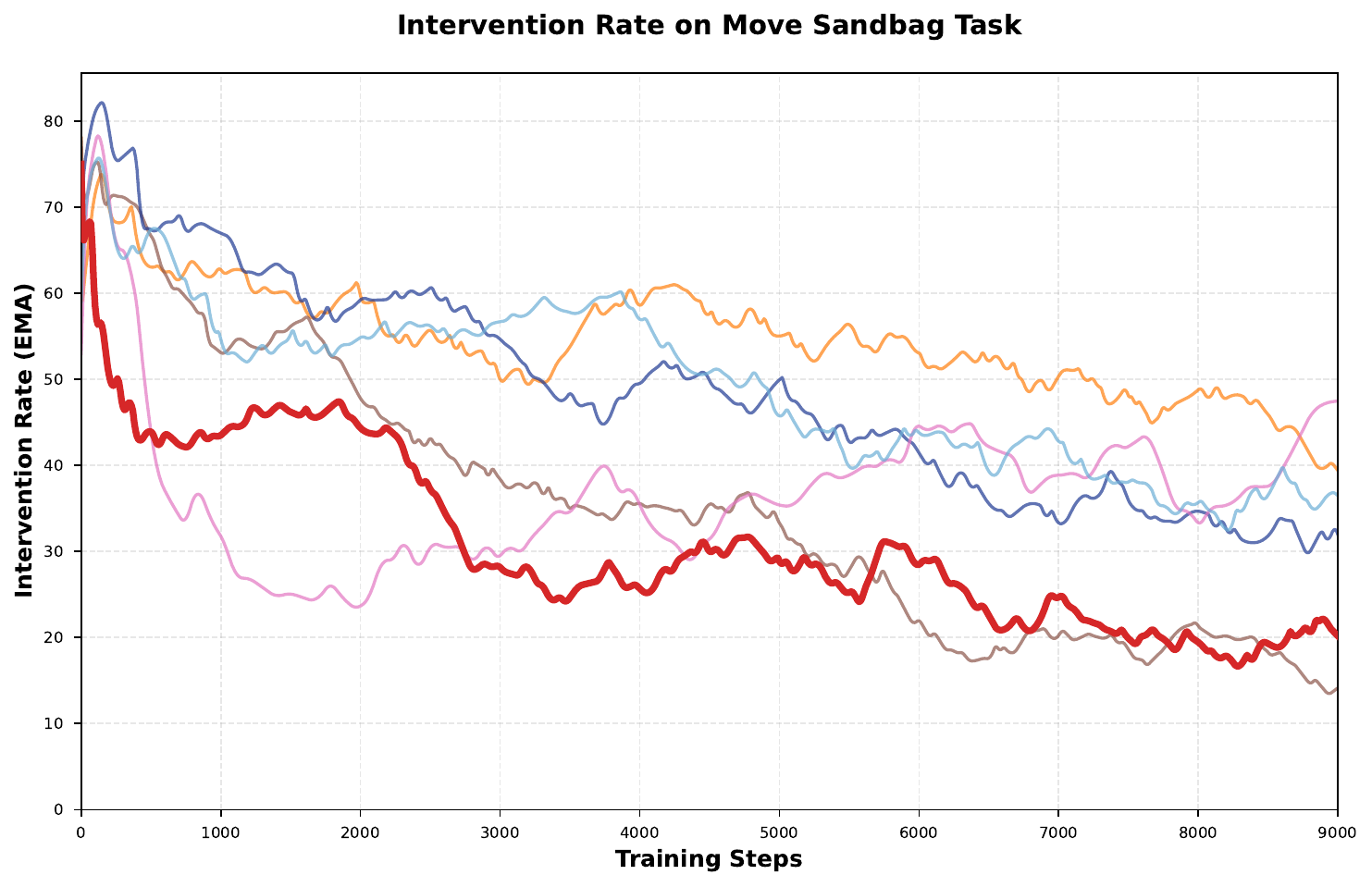}
        \caption*{(b) Intervention Rate}
    \end{minipage}
    \hfill
    \begin{minipage}[b]{0.32\textwidth}
        \centering
        \includegraphics[width=\textwidth]{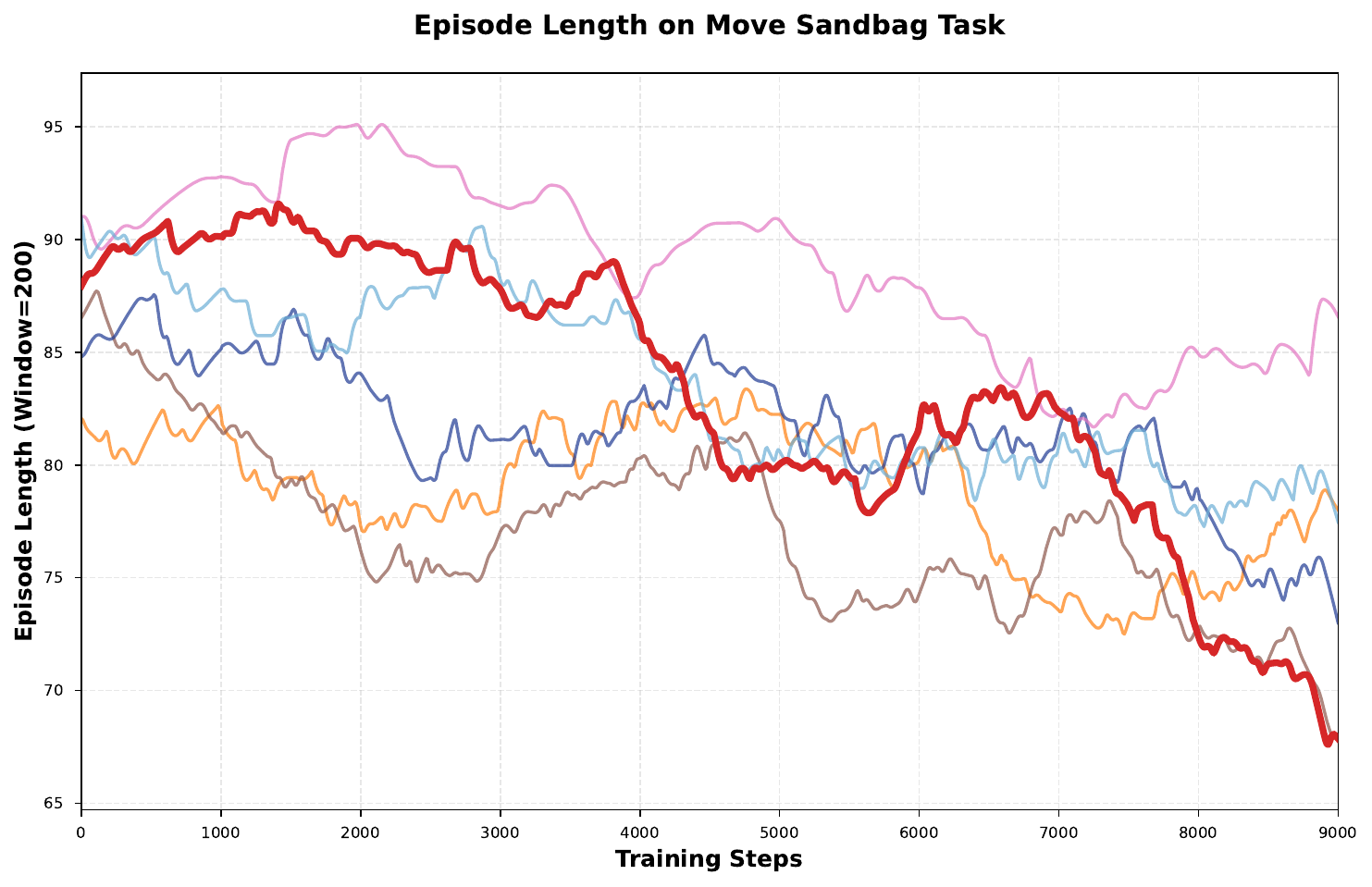}
        \caption*{(c) Episode Length}
    \end{minipage}

    \includegraphics[width=0.8\textwidth]{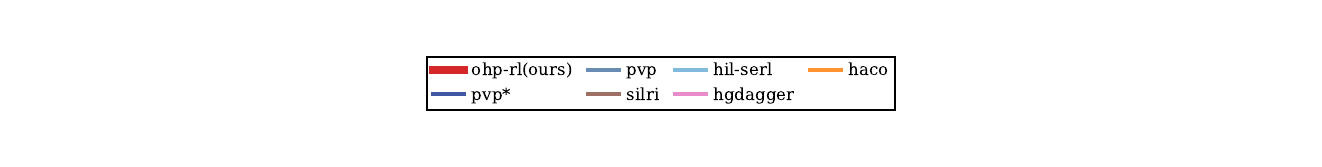}

    \caption{Training performance on the \textit{Move Sand Bag} task. }
    \label{fig:results_sandbag}
\end{figure*}

\subsection{Main Results}

We evaluate OHP-RL from two perspectives: online training dynamics and final autonomous execution without human intervention. During training, we track three metrics: \textit{success rate}, \textit{intervention rate}, and \textit{episode length}. Success rate is computed using a rolling window, intervention rate is smoothed with an exponential moving average, and episode length reflects trajectory efficiency.

\paragraph{\textbf{Press Bell}}
As shown in Fig.~\ref{fig:results_press_bell}, OHP-RL learns the task faster than the baselines while requiring fewer human interventions throughout training. The steadily decreasing intervention rate suggests that control is progressively transferred from the human to the policy without sacrificing task success. OHP-RL also maintains competitive episode lengths, indicating that the learned policy is both reliable and efficient.

\paragraph{\textbf{Push Ball}}
The \textit{Push Ball} task is more challenging because long-horizon exploration is highly sensitive to contact geometry and real-world friction. As shown in Fig.~\ref{fig:results_push_ball}, OHP-RL achieves consistently strong training performance, with higher success rates and a clear reduction in intervention frequency over time. Compared with the baselines, OHP-RL shows better learning efficiency in this harder sparse-reward setting, suggesting that adaptive preference guidance is particularly useful when naive exploration is unreliable.

\paragraph{\textbf{Move Sand Bag}}
The \textit{Move Sand Bag} task is the most challenging task in our benchmark due to randomized initial conditions, deformable-object grasping, and variable placement outcomes. From the training curves, OHP-RL shows the most favorable overall trade-off between task completion and human assistance: while some baselines achieve competitive success rates during parts of training, OHP-RL improves more steadily in the later stage, with substantially reduced intervention frequency and shorter episode lengths toward the end of training. The results of the experiment suggest that OHP-RL learns a more autonomous and efficient manipulation strategy in this challenging environment of deformable objects. Consistent with this trend, Table~\ref{tab:inference_results} shows that OHP-RL achieves the best final autonomous performance on this task.

\paragraph{\textbf{Inference Evaluation}}
We further evaluate each method without human intervention and report the results in Table~\ref{tab:inference_results}. On the simpler \textit{Press Bell} task, OHP-RL matches the best success rate while remaining close to the fastest execution time. On the more challenging \textit{Push Ball} and \textit{Move Sand Bag} tasks, OHP-RL substantially outperforms all baselines in success rate, at the cost of a modest increase in execution time over the fastest baseline, indicating a preference for reliable completion over aggressive but brittle behavior. Overall, the human-in-the-loop baselines remain competitive on the simpler task but degrade markedly under more complex real-world dynamics. HACO performs poorly overall, suggesting that discarding the environment reward is inadequate in our sparse-reward setting, while PVP* also underperforms OHP-RL, indicating that the propagation of intervention comparisons through the critic alone is insufficient. Overall, OHP-RL transfers more effectively to autonomous execution.
\begin{table*}[t]
\caption{Inference performance comparison on real-world manipulation tasks without human intervention. We report final success rate and average execution time.}
\label{tab:inference_results}
\centering
\begin{tabularx}{\textwidth}{l l @{\extracolsep{\fill}} cccccc}
\toprule
\textbf{Task} & \textbf{Metric} & \textbf{HIL-SERL} & \textbf{SILRI} & \textbf{HG-DAgger} & \textbf{HACO} & \textbf{PVP*} & \textbf{OHP-RL (Ours)} \\
\midrule
\multirow{2}{*}{Press Bell} 
& Success Rate (\%) & 98 & 100 & 64 & 0 & 100 & \textbf{100} \\
& Avg. Time (s)     & 2.808 & 3.701 & 7.221 & -- & \textbf{2.373} & 2.435 \\
\midrule
\multirow{2}{*}{Push Ball}  
& Success Rate (\%) & 36 & 46 & 26 & 6 & 32 & \textbf{70} \\
& Avg. Time (s)     & 5.523 & \textbf{4.452} & 10.575 & 11.162 & 6.103 & 6.268 \\
\midrule
\multirow{2}{*}{Move Sand Bag} 
& Success Rate (\%) & 16 & 66& 12 & 4 & 34 & \textbf{68} \\
& Avg. Time (s)     & 7.407 & \textbf{7.335} & 14.275 & 16.235 & 9.596 & 7.760 \\
\bottomrule
\end{tabularx}
\end{table*}

\subsection{Ablation Study}
\label{ablation}

We conduct two ablation studies to better understand the design choices in OHP-RL. The first examines the role of the adaptive preference target on the \textit{Press Bell} task, and the second studies sensitivity to intervention target design on the \textit{Move Sand Bag} task.

\subsubsection{Ablation on Adaptive Preference Target}

To analyze the role of the preference gate and the proposed adaptive target design, we compare OHP-RL against three variants on the \textit{Press Bell} task:
\begin{enumerate}
    \item \textit{Fixed} ($\beta=0.5$): a constant, non-adaptive preference weight.
    \item \textit{Off-target} ($\beta_{\text{target}}=0.5$): the gate target in Equation \eqref{eq:preference_gate} is trained toward a constant value rather than the proposed critic-based target.
    \item \textit{Without RL}: preference guidance learning without the RL actor objective.
\end{enumerate}

\begin{center}
\captionsetup{type=table}
\captionof{table}{Ablation results on the \textit{Press Bell} task.}
\label{tab:ablation_results}
\small
\setlength{\tabcolsep}{5pt}
\renewcommand{\arraystretch}{1.03}
\begin{tabular}{lcc}
\toprule
\textbf{Variant} & \textbf{Success Rate} (\%) & \textbf{Avg. Time} (s) \\
\midrule
Fixed ($\beta=0.5$) & 76 & 2.404 \\
Without RL & 34 & 7.049 \\
Off-target ($\beta_{\text{target}}=0.5$) & 100 & 3.049 \\
\textbf{OHP-RL (Ours)} & \textbf{100} & \textbf{2.435} \\
\bottomrule
\end{tabular}
\end{center}

As shown in Table~\ref{tab:ablation_results}, the \textit{Without RL} variant performs poorly, achieving only 34\% success, which indicates that preference supervision alone is insufficient without reinforcement learning from the environment reward. The \textit{Fixed} variant reaches only 76\% success, showing that a constant preference coefficient cannot adequately balance human guidance and autonomous exploration across states. Although the average learned $\beta$ in OHP-RL is close to $0.5$, these results suggest that its state-dependent variation is essential. Both \textit{Off-target} and OHP-RL achieve 100\% success, but OHP-RL completes the task more efficiently , indicating that the adaptive target based on $A(s_t)$ improves not only final success but also policy quality.

\begin{figure}[!t]
    \centering
    \includegraphics[width=\columnwidth]{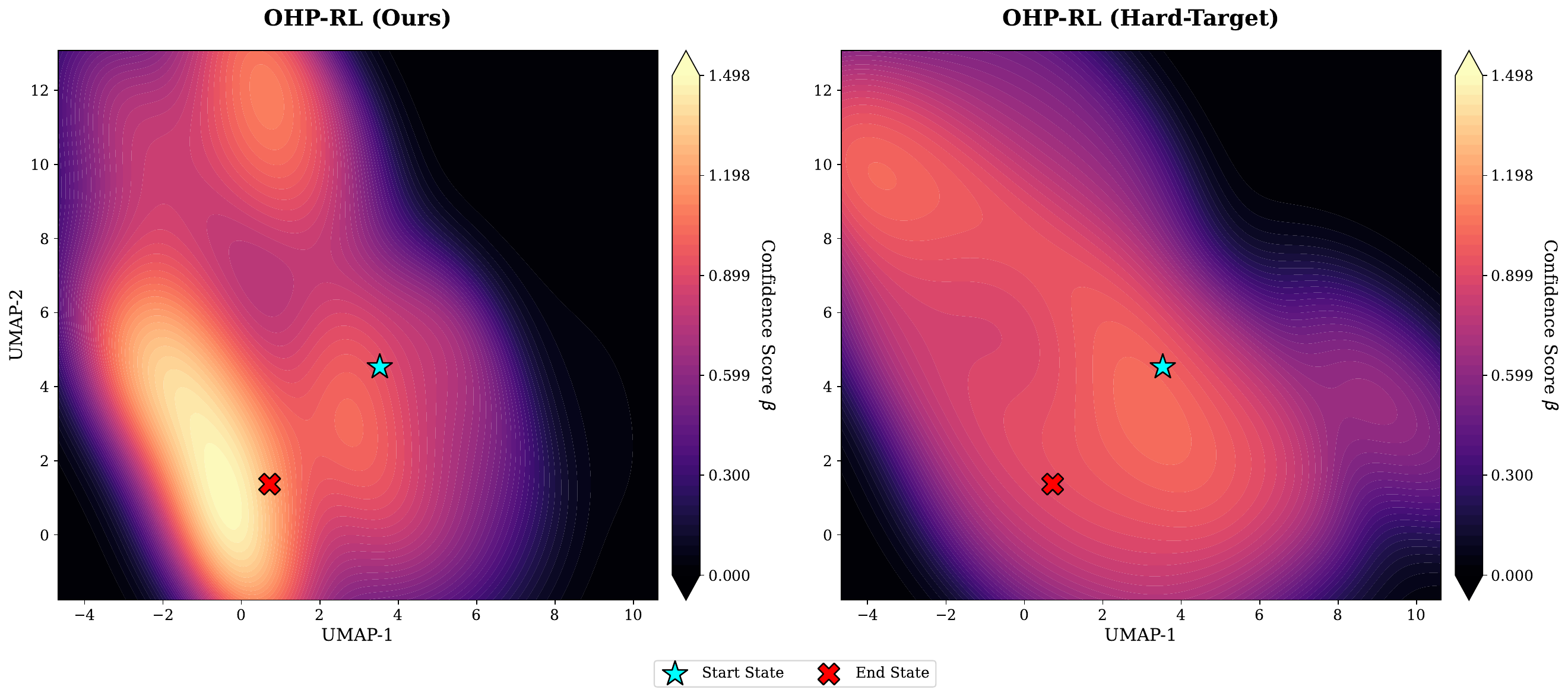}
    \caption{Visualization of learned $\beta$ fields on the \textit{Press Bell} task. The proposed adaptive target focuses high $\beta$ values strictly within hazardous critical zones, whereas the constant target design produces a widespread activation field.}
    \label{fig:ablation_kde}
\end{figure}

Fig.~\ref{fig:ablation_kde} provides a qualitative explanation for this difference. Using $A(s_t)$ to construct $\beta_{\text{target}}$ concentrates large $\beta$ values at the near-danger region, where precise top-down contact is required and side-entry failures are most likely. In contrast, the constant-target design produces a much more diffuse field and fails to distinguish critical states from non-critical ones. This supports the core design of OHP-RL: the advantage-based target allocates preference strength selectively to bottleneck states where human guidance is most useful.

\subsubsection{Ablation on Intervention Target Design}
\label{ablation_safe_region}

We further study how different methods depend on the target of human intervention in the \textit{Move Sand Bag} task. In the main setting, interventions move the robot to a state near the desired trajectory. In the safe-region (SR) variant, all interventions are instead redirected away from current unsafe regions.

\begin{table}[htbp] 
\centering 
\caption{Performance comparison on the Move Sand Bag.} 
\label{tab:performance_results}
\small
\setlength{\tabcolsep}{6pt}
\renewcommand{\arraystretch}{1.05}
\begin{tabular}{lcc}
\toprule
\textbf{Method} & \textbf{Succ. (\%)} & \textbf{Avg. Time} (s) \\
\midrule
SILRI & 66 & 7.335 \\
SILRI + SR & 30 & 7.370 \\
OHP-RL (Ours) & 68 & 7.760 \\
OHP-RL + SR & 52 & 8.748 \\
\bottomrule
\end{tabular}
\end{table}

Under the main intervention design, SILRI and OHP-RL achieve almost the same success rate. However, when all interventions are redirected to a safe region, SILRI drops sharply to 30\%, whereas OHP-RL decreases more moderately to 52\%. This suggests that SILRI is more sensitive to the exact intervention target, which is consistent with its reliance on a behavior-cloning style objective over intervention actions. In contrast, OHP-RL remains more robust under the same perturbation, indicating that the proposed preference guidance formulation can better exploit coarse human interventions without requiring them to match an optimal corrective action exactly.

\begin{figure}[!t]
   \centering
    \includegraphics[width=0.83\columnwidth]{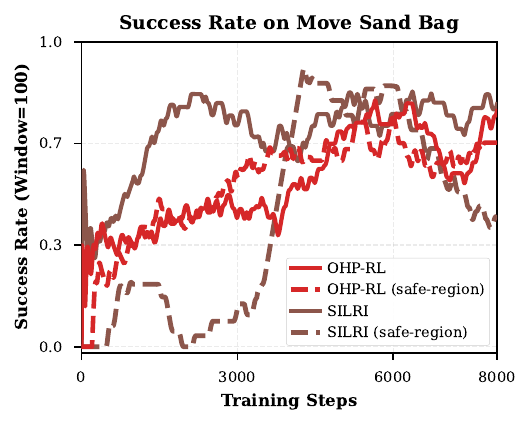}
   \caption{Ablation on intervention target design in the \textit{Move Sand Bag} task. OHP-RL and SILRI perform similarly under the main setting, while OHP-RL is more robust when interventions are redirected to a safe region.}
   \label{fig:ablation_safe_region}
\end{figure}
Fig.~\ref{fig:ablation_safe_region} further reveals a clear difference in training dynamics. Under safe-region intervention, SILRI shows highly unstable learning behavior: its success rate remains low for a prolonged early stage, then rises sharply, and later degrades again. In contrast, OHP-RL maintains a much more consistent improvement trend throughout training, even when the intervention target is no longer task-progress-consistent. This suggests that OHP-RL is less sensitive to the precise form of intervention targets and can exploit coarse human guidance in a more stable manner.

\section{Conclusion and Limitation}
We presented OHP-RL, a human-in-the-loop reinforcement learning framework that treats human interventions as online preference signals rather than only as demonstrations or replay data. By introducing a preference gate conditioned on the state, OHP-RL adaptively regulates when preference information derived from interventions should influence policy learning, enabling the agent to benefit from intermittent and imperfect human feedback while preserving autonomous reinforcement learning under sparse rewards.

Across three real-world manipulation tasks on a Franka robot, OHP-RL achieves strong training performance, reduced intervention effort, and more effective transfer to autonomous execution. The ablation studies further support the two central design choices of the method: the adaptive target improves efficiency and policy quality, while the preference guidance formulation is more robust than imitation-style supervision when intervention targets become coarse or imprecise.

A limitation of the present study is that learning becomes harder as task horizon increases. Long-horizon manipulation requires sustained correct behavior over extended sequences of uncertain interactions, making efficient improvement more difficult in real-world settings. In addition, real-world stochasticity can significantly affect value estimation in sparse-reward tasks. Empirically, tasks with more clearly identifiable success states in observation tend to achieve higher final success rates, whereas tasks with more ambiguous or variable success outcomes are harder to optimize reliably. These observations highlight the need for future work on more robust long-horizon credit assignment and value learning under real-world uncertainty.

\newpage

\vfill

\end{document}